\def\year{2021}\relax
\documentclass[letterpaper]{article} 
\usepackage{aaai21noc}  
\usepackage{times}  
\usepackage{helvet} 
\usepackage{courier}  
\usepackage[hyphens]{url}  
\usepackage{graphicx} 
  \usepackage{soul, color}
  \sethlcolor{green}
  \usepackage{multirow}
\usepackage{natbib}  
\usepackage{caption} 
\newtheorem{example}{Example}

\title{AAAI Press Formatting Instructions \\for Authors Using \LaTeX{} --- A Guide }
\title{Bias in ontologies -- a preliminary assessment}
\author {
    C. Maria Keet\\
}
\affiliations {
    Department of Computer Science \\University of Cape Town, South Africa\\
    mkeet@cs.uct.ac.za
}
\begin{document}

\maketitle

\begin{abstract}
Logical theories in the form of ontologies and similar artefacts in computing and IT are used for structuring, annotating, and querying data, among others, and therewith influence data analytics regarding what is fed into the algorithms. Algorithmic bias is a well-known notion, 
but what does bias mean in the context of ontologies that provide a structuring mechanism for an algorithm's input? What are the sources of bias there and how would they manifest themselves in ontologies?
We examine and enumerate types of bias relevant for ontologies, and whether they are explicit or  implicit. These eight types are illustrated with examples from extant production-level ontologies and samples from the literature.
We then assessed three concurrently developed COVID-19 ontologies on bias and detected different subsets of types of bias in each one, to a greater or lesser extent.
This first characterisation aims contribute to a sensitisation of ethical aspects of ontologies primarily regarding representation of information and knowledge.
\end{abstract}

\section{Introduction}


Bias in models is a well-known topic, which has been popularised to the public with a catchy term ``weapons of math destruction'' \cite{ONeil16}. Nearly all investigations on `models' concern {\em statistical} models created from Big Data by means of knowledge discovery, machine learning, and deep learning techniques.  There are many more types of models, however. The other main category of models within Artificial Intelligence (AI) are {\em ontologies}, which are staple in the knowledge representation and reasoning side of AI. Informally, an ontology is a logical theory of a subject domain, capturing its classes, relations, and constraints that hold among them, which are used for tasks such as data integration, information retrieval, electronic health records, and e-learning \cite{Keet18oebook}. For instance, one may have multiple databases that have to be merged due to a company take-over and one needs to know whether some entity type {\sf\small Customer} or {\sf\small COVID19-Patient} in database$_1$ has the same meaning as {\sf\small Custm} or or {\sf\small COVIDPatient} in database$_2$, respectively, and if it is, a way to declare that, or, e.g., to define precisely what {\sf\small COVID-19 death} means in the mortality statistic. Ontologies can help with it by providing an application-independent representation of the subject domain as a common vocabulary and unambiguous specification of the intended meaning. Besides integration, one also can choose an ontology upfront and use that across applications, such as an electronic patient record system with a medical terminology for classifying or annotating patients's symptoms, disorders  and a treatment that is shared with the insurer; SNOMED and the ICD-10 are popular for that. An example of their Web-scale use is Google's Knowledge Graph that drives search and the creation and maintenance of 
its infoboxes. The one who builds and controls the graph, then, is the one who has the power to control presentation and access to information and possibly also the recording of information, and, as \cite{Vang13} argues in case of Google's Graph, ``to some degree contests the autonomy of the user''. 

We illustrate the general idea of possible issues in the next example with ontology-mediated artificial moral agents.

\begin{example}
The Genet ontology aims to provide a framework to represent multiple ethical theories such as utilitarianism and divine command theory \cite{RK20} so that one can tailor the actions of a robot to the moral preferences of its owner or enhance argumentation in multi-agent systems \cite{Liao19}. A section of its version 1 is shown in Fig.~\ref{fig:genetpart} in black-and-white informally on the left and a selection of the axioms in Description Logics (DL) notation \cite{Baader08} on the right. That the ontology admitted four distinct entities of moral value, rather than just humans, is already an ideaological statement and therewith a bias. 

Now assume that you want to expand the moral circle beyond those four in the ontology, with {\sf\small Robot}. By design, you cannot unless you have the rights and the technology to change it. Let's assume you have those. 

\begin{figure*}[t]
\centering
\includegraphics[width=0.9\textwidth]{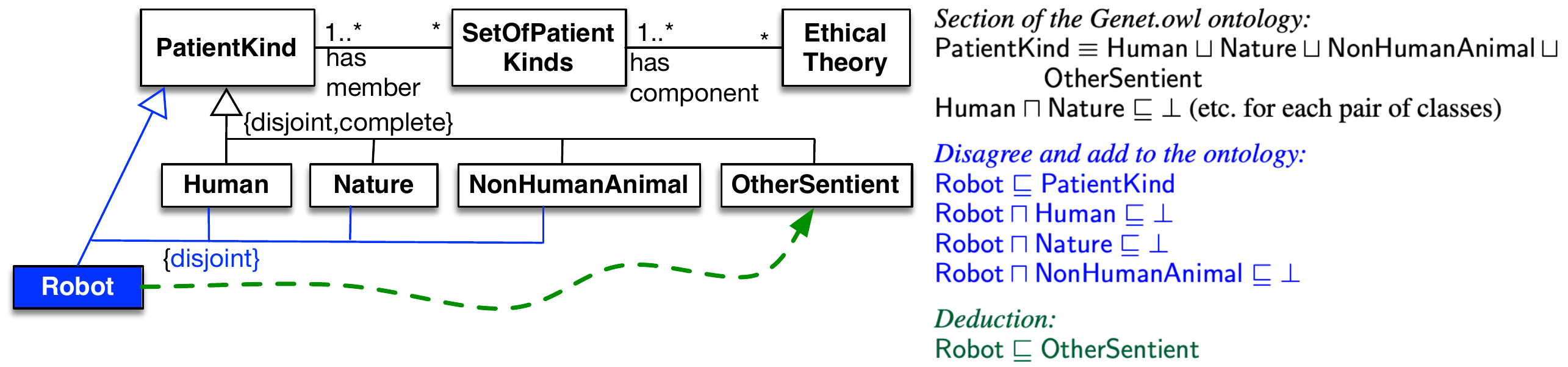} 
\caption{Small section of the OWL version of the Genet model of \cite{RK20} (in black-and-white), a hypothetical addition with {\sf\small Robot} as entity of possible moral value (in blue, solid lines bottom-left), and the deduction (in green, dashed arrow). On the right, a selection of the relevant axioms in DL notation.}
\label{fig:genetpart}
\end{figure*}

There are three options. First, you add  {\sf\small Robot} as a  {\sf\small PatientKind} and since you are sure robots are neither humans, nor nature, nor non-human animals, add those disjointness axioms. It will deduce\\ 
\indent {\sf\small Robot $\sqsubseteq$ OtherSentient}\\ 
regardless whether you wanted that or not. If not---perhaps because you are religiously convinced inanimate objects cannot be sentient---then, second, you could add that they are distinct as well:\\ 
\indent {\sf\small Robot $\sqcap$ OtherSentient $\sqsubseteq \bot$}\\ 
but then the reasoner will deduce\\ 
\indent {\sf\small Robot $\sqsubseteq \bot$}\\
 i.e., the class is unsatisfiable (cannot have instances). The third option is to modify the original axioms and losing compatibility with Genet; e.g., to remove some disjointness axioms or change the completeness axiom.
\end{example}

Ontologies in computing and IT have been popularised since the mid 1990s, with as a major success story the Gene Ontology \cite{GO00} as ontology and the OWL language as the W3C standard \cite{OWL2rec} to represent ontologies in. The popular ontology repository for bio-ontologies BioPortal lists 831 ontologies and the repository of repositories OntoHub claims to have indexed 22460 ontologies of 139 repositories\footnote{figures from \url{https://ontohub.org/}; last checked on 13-1-2021.}. Regarding possible bias in ontologies, aside from ``encoding bias'' \cite{Uschold96} that refers to different formalisations of the same thing, there are few articles. An early paper discusses it in context of the ``Dirty War Index'' tool that claimed to aim to inform public heath in armed conflict settings, which had several biases, such as including ex-army in the civilian group whereas the primary source database did not \cite{Keet09pcr}. \cite{Gomes20} assessed the FOAF terminology through the lens of discursive semiotics as a method. This aimed to mean one has to consider ``the concretization, in language, of a particular social, historical, ideological, and environmental context'', using the specific framework with the ``Generative Trajectory of Meaning'' of Greimas and Court\'es\footnote{referenced as ``Greimas, A. J. and J. Court\'es. 2013. Dicion\'ario de Semi\'otica. S\~ao Paulo: Contexto.''}. The bias analysis, however, was limited to a few well-known ones, being first \& last name vs given \& family name, gender, and the meaning of document. While valid, bias and their causes are more intricate and varied than these. For instance, consider religion, which may be a specialisation of  their ``ideological'', as was the case of the issue of inclusion of homosexuality in the classification of mental disorders\footnote{for a brief overview of its history, see \url{https://en.wikipedia.org/wiki/Homosexuality_in_DSM}} in the United States until DSM-III in 1987. What their approach cannot capture, but is certainly an issue for declarative models, is, among others, the menopausal hormone therapy case: there were at least {\em economic} incentives that determined which attributes ended up in the model with what threshold values in order to classify who is eligible for treatment\footnote{In essence, they narrowed the range of natural variability of concentrations of key molecules to increase the number of women who would be `abnormal' and therewith qualifying for medication, which unintentionally led to an increase in cancer incidence.}.

In this paper, we aim to contribute to systematising the sort of bias that can enter or be present in ontologies and similar artefacts, such as conceptual data models and thesauri. We will seek to provide a preliminary answer to what bias means for ontologies, what their sources or causes are, and how that manifests itself in ontologies. The identified biases types are structured along three categories: high-level philosophical ones, scope or purpose, and subject domain issues. Some of these biases are intentional biases that insiders know very well, but outsiders and newcomers may have to be notified of. For the unintentional biases that can creep in, this will be harder to manage; we do not aim to solve that here, but first inventarise them. Second, we assess a set of COVID-19 ontologies on these biases. These ontologies are under active development, competing, and merging, and highly relevant for data management of the pandemic. The assessment showed that none is free of bias.

The remainder of this paper is structured as follows. We first systematise and illustrate the principal sources, to continue with the COVID-19 ontologies assessment. We then discuss the outcomes and touch upon automated reasoning, and close with conclusions.

\section{Principal sources of bias in ontologies}

Of most interest practically ethically, is the bias with respect to the subject domain. To be able to discuss it properly, we first need to note and `set aside' the straightforward ones of philosophical and engineering (encoding) bias. A summary of the resultant eight types, or sources, of bias is included in Table~\ref{tab:overview}.

\subsection{High-level philosophical issues}

\begin{figure}[t]
\centering
\includegraphics[width=0.45\textwidth]{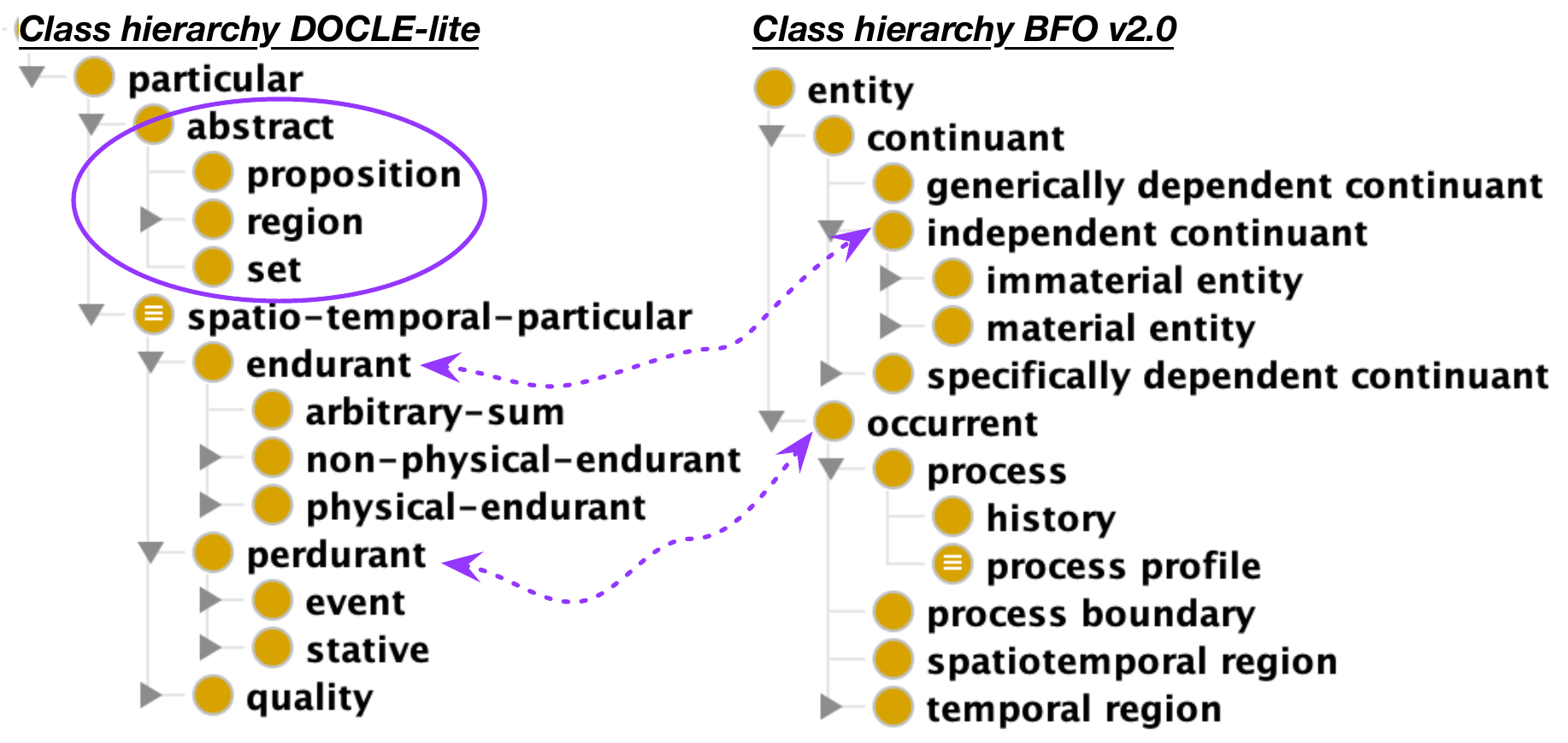} 
\caption{Illustration of some philosophical differences between foundational ontologies: {\tt DOLCE-Lite.owl} has {\sf\small Abstract}, but {\tt bfo20.owl} does not due to its realist stance, and while {\sf\small perdurant} and {\sf\small occurrent} roughly align (dashed arrow), their respective subclasses do not, admitting to different types of perdurant in existence.}
\label{fig:FOdiff}
\end{figure}

Ontologies as an engineering version of the original idea of Ontology by philosophers, and its branch of analytic philosophy in particular.  Most subject domain ontology developers may not care much about the finer distinctions of core notions, but they are there. Practically, for domain ontology development, one would choose a particular foundational or top-level ontology that provides the main types of entities and relations so as to help structuring the content. There are multiple such foundational ontologies in active use, such as BFO, DOLCE, UFO, SUMO, and YAMATO, which make different commitments. Its developers are mostly clear about that on general principles and how it affects the ontology's content, such as acknowledging the existence of abstract entities \cite{Masolo03} or what the core relations in the world would be \cite{Smith05}. See Fig.~\ref{fig:FOdiff} for an example. While it is not trivial to choose which foundational ontology suits the modeller best, it is a deliberated decision, hence, an upfront explicit bias.

There are related debates on whether what is represented is a representation of reality or merely our understanding thereof, or whether there would even be a reality. This is an old and recurring debate (see, e.g., \cite{Merrill10-1}) that has no resolve that everyone agrees on. For ontology development, the key take-away is whether one aims to be faithful to reality (or our best understanding of it) versus ulterior motives, be it rejecting reality or not caring (`post-truth') or knowingly violating it for whatever reason. These different stances act out at the subject domain level where the bias can have most effect, as we shall see further below, and could result either in an explicit or implicit bias.

\begin{table}[t]
\centering
\resizebox{.98\columnwidth}{!}{
\begin{tabular}{l|l|l}
    {\bf Type} & {\bf Subtype} & {\bf [im/ex]plicit bias} \\ \hline\hline
    Philosophical & - & explicit  \\ \hline
    Purpose & - & explicit \\ \hline
    \multirow{6}{*}{Subject domain} & Science & explicit  \\
     & Granularity & either  \\
     & Linguistic & either  \\
     & Socio-cultural & either  \\
     & Political or religious & either  \\
     & Economics & explicit  \\
\end{tabular}
}
\caption{Summary of typical possible biases in ontologies grouped by type, with an indication whether such biases would be explicit choices or whether they may creep in unintentionally.}
\label{tab:overview}
\end{table}

\subsection{Scope or purpose}

In theory, ontologies are supposed to be application-independent, so as to be a solution to the data integration problem; if they are tailored to the application nonetheless, they may become part of the problem. In praxis, this application independence may not always hold. Developing an ontology for the sake of it may be an interesting endeavour, but someone has to fund it and it helps to have a use case scenario to motivate for the development of it.  This may affect what is represented and how and is, or at least should have been, an explicitly stated bias motivated by pragmatics, if it can be considered a bias---as \cite{Uschold96} do---since they are engineering choices rather than bias on the knowledge itself.

\begin{figure}[t]
\centering
\includegraphics[width=0.42\textwidth]{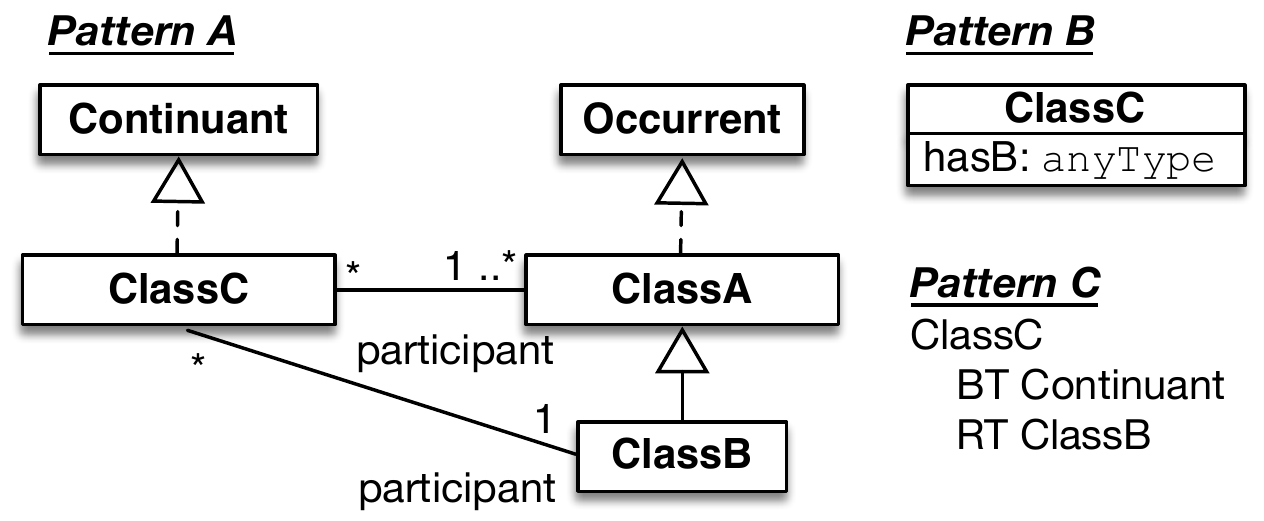} 
\caption{Three different patters with a purpose bias: Pattern A is biased toward a scientific approach with increasing precision (and a bias toward 3-dimensionalism philosophically), Pattern B indicates a conceptual data modelling influence or purpose, and Pattern C takes a thesaurus-like approach useful for document annotation.}
\label{fig:patterns}
\end{figure}

Three patterns of representation for different purposes are shown in Fig.~\ref{fig:patterns}, summarising common encoding biases. To illustrate those, consider the following situation, represented in DL for brevity. Ventilation is an undisputed treatment for COVID-19 patients and is being used, and so let us consider three options:
\begin{itemize}
\item If the scope or purpose is the be as detailed and reusable as possible, and knowing that  ${\sf\small Treatment}$ is a perdurant in philosophical terms, operating within the 3-dimensionalism bias, then\\ 
\indent \hspace{2mm} ${\sf\small Ventilation \sqsubseteq Treatment}$\\ 
is the bare minimum to declare, and availing of the core relation of {\em participation}, then also\\ 
\indent \hspace{2mm} ${\sf\small HospitalisedPatient \sqsubseteq \exists participatesIn.Treatment}$\\ 
One then could assert that our hospitalised COVID-19 patients participate in the ventilation treatment. 
\item A compact representation of the same state of affairs results in faster data processing; e.g.,\\ 
\indent \hspace{2mm} ${\sf\small Patient \sqsubseteq \exists isOnVentilator.}{\tt Boolean}$\\ 
so as to use the ontology language to develop a conceptual data model for database development, rather than the traditional EER language for relational databases. 
\item Another purpose could be annotation of literature to better manage it. Then neither the boolean nor all those constraints and relations are needed, but it would focus on casting the net wide on terminology with one preferred and several alternative labels including, but not limited to, {\sf\small ventilator support}, {\sf\small ventilation therapy}, {\sf\small mechanical ventilation}, and {\sf\small invasive ventilation} with BT {\sf\small Ventilation} and RT {\sf\small Patient}. 
\end{itemize}
The ontologist may complain about the latter two options as woefully underspecified, which is their bias, whereas the two tool developers may complain that the first option is needlessly complicated due to their bias for simplicity.

\subsection{Subject domain}

The list of bias sources described in this section  overlaps with Gomes and Bragato Barros' one \cite{Gomes20}, but is extended with three categories, including one from \cite{Keet09pcr}. In addition, we indicate whether they concern mainly intended or unintended biases, or both, and illustrate each one in order to demonstrate relevance. \\

\textit{Difference of opinion on reality and science.} Even under the assumption of a commitment to the existence of reality, one still could disagree. A common example is whether a virus is an organism or not---it is not by any extant definition of what an organism is---and even bio-ontologies and medical terminologies do not agree; compare, e.g., the CIDO \cite{He20} versus the SIO \cite{Dumontier14} and the NCI Thesaurus\footnote{\url{https://ncit.nci.nih.gov/ncitbrowser/ConceptReport.jsp?dictionary=NCI_Thesaurus&ns=ncit&code=C14283}; 18-1-2021.}. 
More broadly, it concerns either the insufficient insight or competing theories that the scientists still have to investigate, of which it is assumed that eventually there will be an agreement, or there are delays in propagating discoveries into the ontologies. In other fields of research, there are inherently competing theories, such as capitalism and socialism, that would result in a different domain ontology of economy. They are all intended choices and biases.\\

\textit{Required or chosen level of precision/granularity}. It is a general question in ontology development how detailed it should be and how deep the taxonomy should go. Less detail therefore may be an {\em act of omission}, an indication of `not needed', or a `ran out of time' to be included for a next version. Inspecting an ontology in isolation, this is impossible to determine unless either of the latter two are indicated in the annotations. For instance, The Gene Ontology has three versions: a GO basic that excludes several relations between entities, the GO, and a GO plus with additional axioms\footnote{\url{http://geneontology.org/docs/download-ontology/}; last accessed 13 January 2021.}. 

An act of omission is aforementioned aggregation of ex-military persons with, say, non-involved persons as one group of {\sf\small Civilians}: the source had a more detailed categorisation that was abstracted away so that it resulted in one party (of the authors' side) shown in a more favourable light  \cite{Keet09pcr}. Similar issues exist for other conflict databases, which may be intentional or unintentional. For instance, a bombing target may be recorded as an instance of having targeted a {\sf\small Government building} if that is the only category available, or more precisely if it had any subclasses, such as, say, {\sf\small State hospital}, {\sf\small State medicine manufacture plant}, {\sf\small Military base}, and {\sf\small Homeland security torture bunker}, and more, rather than one layer of subclasses as in \cite{Veerasamy12}. Similarly, one could have one aggregate group, say, {\sf\small Foreign National} with as alt-label {\sf\small Alien} for the USA, or also include subclasses such as {\sf\small Migrant} and {\sf\small Refugee}, and further subclasses such as {\sf\small Economic migrant}, {\sf\small Spousal migrant}, {\sf\small Critical skills migrant}, and so on. Such differentiations, or absence thereof, may be intended or they may be unintended and even change over time when the subject changes, such as new immigration policies and different ways of conducting conflict (e.g., cyber attacks rather than bombings).\\

\textit{Cultural-linguistic motivations}. Anyone who has learned a second language has come across untranslatable words or at least fine semantic distinctions.  The question then arises if, and if so, when, a difference ends up as a bias in the ontology or not. For instance, English has only one term for river---all rivers are just rivers---whereas French makes a distinction between a {\em fleuve} and  a {\em rivi\`ere}---one flows into another river, the other flows into the sea---that somehow has to be represented and the ontologies aligned \cite{McCrae12}, and likewise for observed differences in part-whole relations \cite{KK18fois}. One may argue that in both examples, the reality is the same but they have varying descriptions, or take reality with a grain of salt and state there are different realities depending on language, or that there are different conceptualisations. 

A borderline case between cultural-linguistic preferences and political bias are the false friends, where a term in a language has a different meaning or connotation in different countries where the language is spoken, due to historical differences across countries. For instance, `herd immunity' is a common term in American and British English, but is being rebranded as `population immunity' in South African English, since the former has the connotation of non-human animals that people do not want to be associated with. It is also different in other languages; e.g., in Spanish, it is {\em inmunidad de grupo} and Dutch {\em groepsimmuniteit}, i.e., `group' immunity rather than `herd'. Note that this is distinct from mere synonym confusion, such as a Football Ontology where it is unclear from the name whether it refers to the soccer football or American football or eraser/rubber/condom mixups, and orthographic differences (e.g., color vs colour), which can be accommodated in the ontology with labels and finer-grained language-coding schemes (e.g., {\tt\small @en-uk} and {\tt\small @en-us etc.}).

The chance that a monolingual ontology development team from one cultural identity in one country builds in such a bias is substantial, and it can be reduced by constituting a more diverse team of ontology developers who at least speak several languages among them. Any bias built in may be intentional or unintentional. For instance, McCrae's team \cite{McCrae12} was very multilingual and so it was easy to observe the difference and propose a solution. In contrast, if one then develops an ontology afterward knowingly only including the non-differentiating {\sf\small River}, then that is an explicit bias. \\

\textit{Socio-cultural factors}. This concerns hows society is organised, with the assumptions that underlie it and history how it came about, and practical effects it may have when developing the ontology. This may be organisational structures, who lives with whom, demographics, allocation of resources, or social geography that influences what is salient and what not. 

For instance, who can marry whom and how many is a well-known point of variation across the world, which can cause difficulties for multinational organisations to harmonise that in one system. For instance, it may be a company policy that one can insure the spouse of the employee, requiring a statement alike\\ 
\indent ${\sf\small Employee \sqsubseteq \forall marriedTo.Spouse}$\\ 
but should the model also include\\ 
\indent ${\sf\small Employee \sqsubseteq\, \leq 1\, marriedTo.Spouse}$\\
 i.e., at most one spouse? Should the gender of the spouse be recorded or {\sf\small marriedTo} be defined as holding between humans and no more? {\em Any} answer will have a bias baked into it. For an ontology to be as general as possible the most permissive combination represented, and any constraints would have to go into the conceptual data model for the specific database. 

A concrete example is the relatively popular GoodRelations Ontology for e-commerce \cite{Hepp08}. It lists several payment methods, such as invoice, cash, and PayPal,  and limits the `on delivery' to cash, but {\em cash-less} options on delivery are just as possible, such as a pre-paid card or QR-code payment option, is missing, which is a non-uncommon mode of payment in areas where robberies are common. Also, its {\sf\small Business} assumes that they are legally registered, which may well hold in Europe where the ontology was developed, but in many other countries there is a vast network of the informal economy that does trade online with their smartphones and it has no specific opening hours. 

Socio-cultural factors may also influence the content of medical terminologies, such as the perception of alcohol use across cultures, in-groups, and age, and what would be considered as having a drinking problem. A recent example is demonstrated by a comparison between the DSM-IV, DSM-V, and ICD-10 medical terminologies on issues with alcohol intake, where the criteria were changed. This resulted in an increase in {\sf\small Alcohol Use Disorder} using DSM-V compared to the DSM-IV criteria (based on the same data), primarily due to lowering the threshold for the number of diagnostic criteria required for it and increasing the number of criteria through replacing one class with four new classes that were arguably features of it \cite{Lundin15}. This change in the  lightweight ontology has been blamed on a combination of socio-cultural factors and scientific disagreement \cite{Wakefield15}. \\

\textit{Political and religious motivations.} The line between societal bias, political, and religious may be difficult to draw depending on the case. Aforementioned DSM, which ought to be based on science, was not entirely and likely was influenced by religious viewpoints at least in some instances. Since the separation between state and church may not be all that separate, it practically may not be possible to disentangle the two. A clear-cut case is where the entity type {\sf\small Aggrieved group}, as a neutral term, enters the ontology as {\sf\small Terrorist organisation} as preferred label; 
concretely, there are  {\sf\small terrorist} and {\sf\small terroristgroup} in the terrorism ontology of \cite{Jindal20}  whereas there is an {\sf\small ActorEntity} with various types of {\sf\small Insider}s and {\sf\small Protestor}s in the Cyberterrorism ontology \cite{Veerasamy12}.  

As with society and language matters, these issues more easily come to light if the team of ontology developers is diverse or at least has diverse knowledge to bring in. Also here, such differences---biases---may be intentional or not.\\

\textit{Economic motivations}. The, perhaps, most well-known arena where economic motivations play a role, is the recognition of something as a disorder or disease, from which follows that it deserves at least funding of a treatment if there is one as well as resources for prevention and research. The Obesity Society's panel of experts even stated this bluntly as the main reason in favour of classifying obesity as a disease \cite{TOS08}. Its recognition is good for big pharma and possibly also the patients, but costly for insurers, which results in tension. For the ontology, it means that it is in our out, where the ontology comes into play in particular in electronic health records (how an observed finding is noted in the record, which treatments are linked to it) and further down the pipeline when the electronic records with their ontology, such as SNOMED CT, are linked to the pharmacy and the insurer's databases. There is a benefit to the data integration if the ontology used for it is grounded on evidence-based medicine and  in one's favour; if it is not, it can be an uphill battle on multiple fronts. 
These issues are well-known and therefore can be classified as intended biases.


\section{Assessment of ontologies: the COVID-19 Ontologies}

To assess the notion of bias in ontologies beyond the concrete selected examples in the previous section, it would support the bias source identification for a set of ontologies in roughly the same subject domain. The reasoning is that since there are several ontologies in that given domain, there must have been a reason to develop more than one rather than to stick with one effort or to combine efforts. This may be due to bias, but not necessarily so. This limits the choices for assessment. Comparisons of foundational ontologies are abound (see \cite{Partridge20} for the most recent attempt) and would have a less clear impact on domain ontologies' possible biases that affect applications that people use. Of the core and domain ontologies, there are a few on time and measurements, many on health and medicine (e.g., 37 are contextualised in \cite{Haendel18}), data mining, organisations and government, and others, which are more or less stable and more or less maintained.

We identified a set of ontologies in a same subject domain, of which the authors have sufficient knowledge about the domain to assess it, and that are under active development and maintenance, so that it has an increased chance of the assessment outcomes to be taken into account. A downside of the latter selection criterion may be that any issues observed may have been resolved in the meantime between assessment and review or publication of this paper. Nonetheless, given the urgency of the theme, we chose to assess the COVID-19 ontologies on bias. The next section will contextualise each ontology briefly and the section thereafter contains the assessment.

\subsection{Ontology descriptions}

The Coronavirus Infectious Disease Ontology (CIDO) \cite{He20} is an ontology that was developed within the overarching OBO Foundry approach \cite{Smith07}: it took a community-based development approach and reuses, among others, the Infections Disease Ontology that in turn is linked to the top-level ontology BFO \cite{Arp15} and therewith adhering to some of its principles of structuring knowledge and philosophical stance of realism. The scope of the ontology was aimed at knowledge and information about the SARS-CoV-2 virus and host taxonomy data, its phenotype,  and drugs and vaccines to foster data integration. 
The CIDO v1.0.109 was used for the assessment, to keep with the time frame where all ontologies were released around July-August 2020, therewith reducing the chance of mutual influence; in particular, the smaller {\tt cido-base.owl} file (downloaded on 20-7-2020) with the relevant imports was assessed, which contains 82 classes, 15 object properties (relations), no data properties (attributes) and one individual, and 90 logical axioms and is within the OWL Full profile due to issues with undeclared annotation properties and a few undeclared classes. Aside from that, logically, it is expressible in $\mathcal{ALEHO}$ in DL terminology, that is, a basic hierarchy with existentially quantified properties and an occasional nominal (instances made into a class). 

The CODO \cite{Dutta20} has as purpose to assist in representing and publishing of COVID-19 data from the disease course perspective and has subject domain scope COVID-19 cases and patient information. That is, it aims to be an  component in IT systems for healthcare, rather than take a medical or research angle. The {\tt CODO V1.2-16July2020.owl} was used for the assessment, which contains 51 classes, 61 object properties, 45 data properties, 56 individuals, and 463 logical axioms. It is within the OWL 2 DL profile, and $\mathcal{SHOIQ}(D)$ more specifically, or: it is an expressive ontology that uses many of the OWL 2 DL constructs available in the language. 

The COVoc, developed by the European Bioinformatics Institute, has as purpose to support navigating and curating the literature on COVID-19, and in particular the scientific research of it; documentation of its rationale is available as a workshop presentation \cite{Pendlington20}. 
Its first, and latest, released version is slightly later than that of CIDO and CODO, although all had their drafts in June 20, which did not affect its contents. The {\tt covoc.owl} was used for the assessment (v d.d. 28-8-2020), which contains 541 classes, 179 object properties, no data properties or individuals, and 672 logical axioms. It is within the OWL Full profile due to a subset property issue with the annotation properties; without that and just the logical theory, it is expressible in $\mathcal{ALCHI}$, consisting of a basic hierarchy with existentially quantified properties and a few subproperties and inverses.

The ``vocabulary for COVID-19  data'', available at \url{http://covid19.squirrel.link/ontology/}, has been excluded, because its contents is different from the other three, in that it is not for COVID-19 data but to label datasets of COVID-19 data, such as {\sf\small Dataset of the Robert Koch-Institut}.

\subsection{Bias assessment}

The presence and absence of the different types of bias is summarised in Table~\ref{tab:covidOntos}, and will be illustrated and discussed in the remainder of this section.




\begin{table}[t]
\centering
\begin{tabular}{l|c|c|cl}
     {\bf Bias} & {\bf CIDO} & {\bf CODO} & {\bf COVoc} \\ \hline\hline
      Philosophical & + & - & +  \\ \hline
      Purpose & - & + & + \\ \hline
      Science & - & - & +  \\
      Granularity & $\pm$ & + & +  \\
      Linguistic & + & - & -  \\
      Socio-cultural & + & + & +  \\
      Political or religious & + & + & +  \\
      Economics & - & - & -  \\
\end{tabular}
\caption{Presence or absence of bias in the three COVID-19 ontologies examined.}
\label{tab:covidOntos}
\end{table}

\subsection{CIDO}

There are two socio-cultural biases in the CIDO. First, there is a {\sf\small COVID-19 diagnosis} class with three subclasses: negative, positive, and presumptive positive. There are two aspects to this: the [disease]-positive/negative labeling, which has clear HIV connotations with all the stigmatisation that comes with it. This may be less prevalent in a country like the USA where the incidence is relatively very low, but in countries where it is endemic, such as South Africa, such labelling can be harmful. It easily could have been, e.g, `infected', `detected', or `present' and `not infected', `absent' or `free'; that said, the positive/negative is a pervasive issues across languages and countries. The third category, `presumptive positive', elicits a negative connotation, plays into people's fears, and would brand people that are statistically unlikely to have it, since many countries aim for at most 5-10\% positivity rate. Neutral, and more accurate, terminology would be, e.g., `pending result', `awaiting test outcome', or `under investigation'. 

A similar bias in the other direction---of unwarranted optimism---is the assumption of\\
\indent {\sf\small COVID-19 experimental drug in clinical trial $\sqsubseteq$\\
\indent\indent COVID-19 drug}\\ 
 noting that\\ 
 \indent {\sf\small COVID-19 drug $\sqsubseteq$ \\
 \indent\indent $\exists$  treatment for.COVID-19 disease process}\\
is asserted in the ontology, and thus entails that {\sf\small COVID-19 experimental drug in clinical trial} {\em is} a drug already and {\em is} being part of regular treatment processes of COVID-19, since the property of {\sf\small $\exists$treatment for.COVID-19 disease process} is inherited down into the hierarchy. This is wishful thinking. A substance under investigation that is being evaluated is not necessarily effective or safe and for it to be a drug, it has to be that and also have been approved by the regulatory body.

A minor language note is {\sf\small drive-thru} instead of {\sf\small drive-through} for testing stations, but this can easily be addressed by providing alternative labels. Other US-centric indications are naming SARS-CoV-2 also the {\sf Wuhan virus}, which was rarely used outside the USA since it was advocated by President Trump and his policies toward China, and {\sf\small FDA EUA-authorized organization} as the only other organisation as sibling of {\sf\small drive-thru COVID-19 testing facility}. The latter may also be an instance of ran-out-of-time, since the authors of the accompanying paper (\cite{He20}) have diverse affiliations. 

Regarding philosophical bias, this is evident by its embedding in the OBO Foundry suite \cite{Smith07}, through its partial reuse of ontologies within that framework, such as OBI and IAO, as well as the organisational principles how the ontology is structured, which follows the BFO foundational ontology design principles \cite{He20}.

In sum, it does try to take the science angle to representing knowledge about COVID-19, but with a few biases toward USA-centrism, which reduces its off-the-shelf potential. Or: if this were to be used in Europe or any of the key Global South countries with ample research, testing, or production capacities, such as India and South Africa, then they would have to modify it first.

\subsection{CODO}

The CODO fares slightly better on the {\sf\small Laboratory test finding}, which can be negative, positive, or {\em pending}, rather, although also here the positive/negative may benefit from a relabeling. Also, it does have the well-known gender issue, captured in the axiom\\ 
\indent {\sf\small Gender type $\equiv$ \{Female, Male\}}\\ 
A clear socio-cultural axiom in the ontology is\\ 
\indent {\sf\small InfectedSpouse $\sqsubseteq$ InfectedFamilyMember}\\ 
\indent {\sf\small InfectedFamilyMember $\sqsubseteq$ Exposure to COVID-19}\\ 
One can argue about omissions or time constraints, since the only family member that can be infected is the spouse according to CODO, but  there may be more family members. The cultural bias here is the concept of the {\em nuclear family} that consists of the parents and their children. Globally more applicable would be to talk of a {\em household}, however that may be composed. This, since there may be live-in grandparents, cousins, nannies, domestic workers, and so on, and spouses may not live together in one household due to being migrant workers. An early example of such complexities in the context of COVID-19 can be found in \cite{Parker20} and if CODO were to be used elsewhere, it would have to revise this branch in the ontology. 

The purpose is indicated through its heavy use of data properties, hence, more alike a model for recording data than for representing the science of COVID-19 or SARS-CoV-2. A substantial amount of information would be usable across countries trying to record data about patients. 
One class is specific to the country of its developers, India, which is the {\sf\small Mild and very mild COVID-19}, which is one of the three categories mandated by its government rather than the modellers' granularity bias, as the authors noted in the annotations of {\sf\small Patient}.  

\subsection{COVoc}

COVoc clearly states that its purpose is COVID-19 scientific literature `triage', and it is informally well-known that knowledge organisation systems for literature annotation is focused in facilitating that rather than being concerned with ontological precision or correctness. Its contents are not clearly structured as a result of this bias, in the sense that there are many top-level terms and mixing of classes and instances, but some aspects, such as the use of the IAO and import of the RO, may indicate some leaning to the OBO Foundry stack as well.  Its actual contents regarding bias raises several questions.

One is of granularity, and perhaps also focus or time, which are straightforward omissions, such as listing only two continents, Asia and Europe (there are 4-7, depending on how one categories),
and a mixture of omission and politics regarding the countries, since there are 10 subclasses of {\sf\small Country}, of which two are disputed (Hong Kong and Taiwan) and one is definitely an error, since West Africa is not a country but a region on the African continent. 

Scientifically, the low-hanging fruit for bias detection is\\
\indent {\sf\small Virus $\sqsubseteq$ Organism}\\
because a virus is not an organism, and that there are several {\em disorders} that are subclasses of {\sf\small Disease}, such as\\ 
\indent {\sf\small headache disorder $\sqsubseteq$ Disease}\\ 
\indent {\sf\small anxiety disorder $\sqsubseteq$ Disease}\\ 
whereas they are distinct medically. With a benefit of the doubt, one might argue they may be layperson commonsense assumptions, but these would then be rather serious ones for an ontology for scientific literature. Further scientific perspectives are built in by recording {\em symptoms}, such as {\sf\small Cough} and {\sf\small Diarrhea} as subclasses of {\sf\small phenotype}, with phenotype defined as ``The detectable outward manifestations of a specific genotype.''. This a very gene-centric view on the body.

Gender is not present, but {\sf\small biological sex} is used instead. The only biological sex recorded in COVoc is {\sf\small male}. Published literature on women and COVID-19 easily dates back to March 2020 (e.g., \cite{Li20}), however, which is well before COVoc's development.

 Since CIDO and CODO had different test statuses, it was examined in COVoc as well. It has seven options: there is a {\sf\small possible case} (meeting clinical criteria), a {\sf\small probable case} (meeting clinical criteria, with epidemiological link, or meeting the diagnostic criteria), and once confirmed there are five types of {\sf\small infection}: asymptomatic, mild, moderate, severe, and critical. There are no test outcomes, only, {\sf\small rapid testing} and {\sf\small serology test}.

Since many terms are plain science terms, like {\sf\small replicase polyprotein 1a (BtCoV)} and {\sf\small cryogenic electron microscopy}, there are no obvious language or linguistic issues in the sense of bias, other than an English bias that nearly all existing ontologies have. One arguably may be COVoc's {\sf\small Social distance} compared to {\em physical distancing}, but the latter has the former as net effect and so the line is not clear. Economic motivations or possible benefits or losses are not evident either.

\section{Discussion: Consequences of bias in ontologies}

Having established that there are indeed biases in ontologies, does it really matter beyond the hypothetical issues and the increased morbidity and mortality in case of the hormone replacement therapy? They do and there are several ways where it can affect it, with the three principle ones being due to {\em omissions}, {\em incorrect attributions}, and {\em undesirable deductions} that are logically correct but not ontologically or not according to the other bias.  

Omissions and incorrect attributions have a direct effect on data analysis, since they increase the amount of noise (technically speaking) when the ontology is used for ontology-based data access and literature annotation and search. For instance, while mortality rates of men are higher for COVID-19, relatively more women get infected; if that cannot be annotated, since absent in COVoc, then the emerging literature is harder to search to find studies on possible causes for why women are tested positive more often than men. Similarly, the lack of the concept of household, or at least more family members, in CODO, prohibits finer-grained recording of the chain of infection 
and thus more likely to lose control of the spread of the virus.

Incorrect attributions have to do with the annotator not finding the desired knowledge in the ontology and then using something else for it. For instance, if, say, Ireland were to use CIDO, then the {\em walk-through} testing facility at Dublin Airport can be approximated by CIDO's {\sf\small drive-thru} one in the sense of passing by or {\sf\small FDA authorised} in the sense of being an official test location. More generally: annotators choose approximations based on different criteria, so any data analysis then will both miss instances and have false positives. Also, and aside from the fact that the different variations on test outcomes contributes to the data integration problem, a {\sf\small presumptive positive} annotation is, on the whole, an incorrect label in about 45-95\% of the time and would seriously distort epidemiological investigations and overload tracking and tracing efforts on top of it. That is, as long as the ontology does not fully characterise all the properties of an entity type so as to be clear on the exact semantics, there is a heavier reliance on the term, with language alone being an easier target to be used or interpreted with bias.

\begin{figure}[t]
\centering
\includegraphics[width=0.45\textwidth]{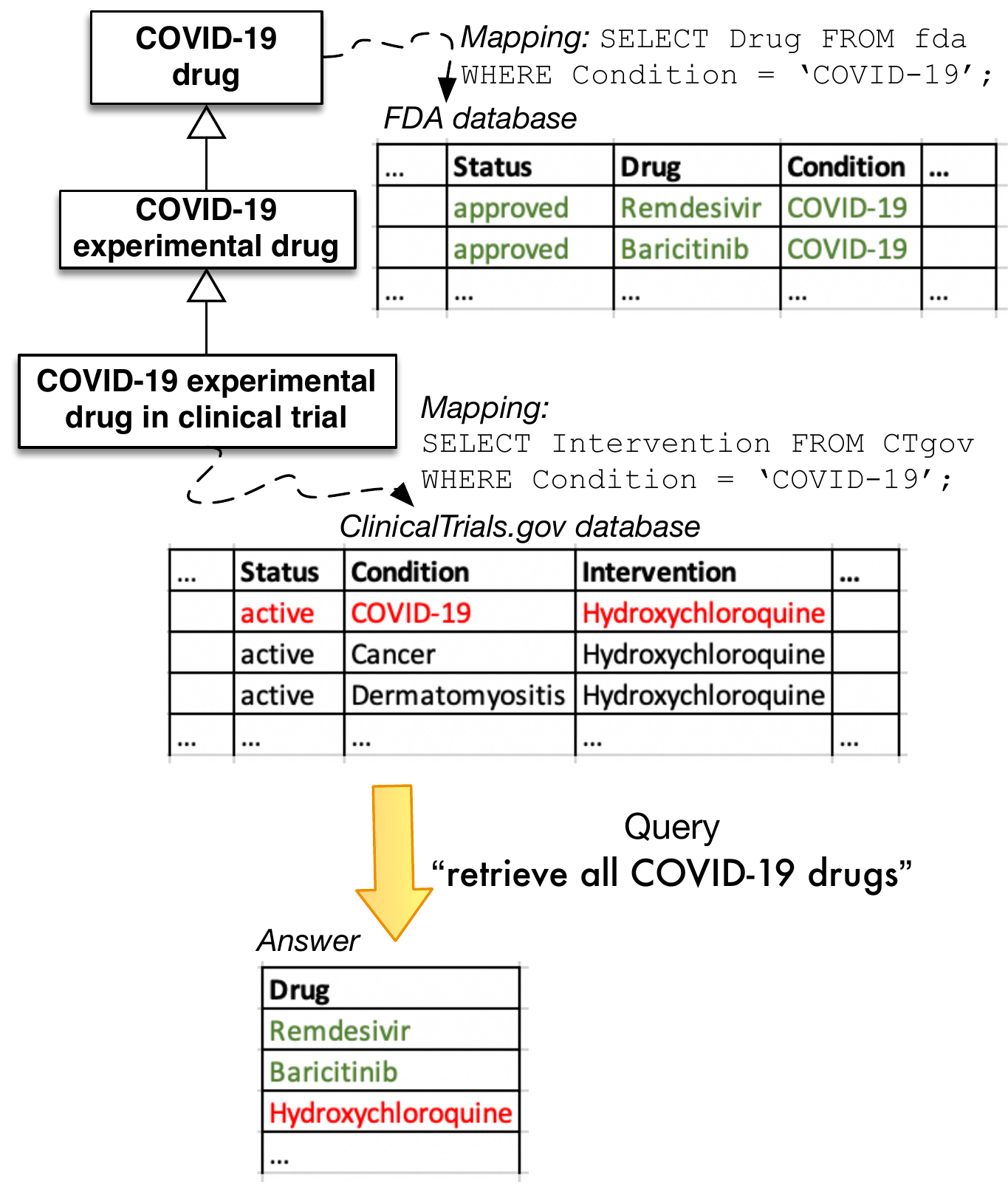} 
\caption{Ontology-based data access and integration scenario with CIDO and two database tables, from the ClinicalTrials.gov and FDA (selection shown, and mappings to the OWL classes are abbreviated). Retrieving {\sf\small COVID-19 drug} recursively fetches from the subclasses in the hierarchy and takes the union of the query answer to each subclass, thus then returning that hydroxychloroquine is already a COVID-19 drug, which is an undesirable deduction from both a scientific and regulatory standpoint.}
\label{fig:obdaEx}
\end{figure}

An example of an {\em undesirable deduction} resulting from a bias built into an ontology would be the drugs with CIDO, which is illustrated in Fig.~\ref{fig:obdaEx}. CIDO aims to facilitate data integration \cite{He20}, which could be done with, say, ontology-based data access (OBDA) and integration to link data to ontologies \cite{Poggi08} where each class and object property in the ontology is mapped to a query over the database(s). A query over the ontology then avails of those mappings to retrieve the answer, together with the knowledge represented in the ontology. Hydroxychloroquine is still used as an experimental drug in COVID-19 clinical trials and is listed as such in the clinical trials registry database\footnote{There are 24 active trials with hydroxychloroquine for COVID-19 at the time of writing (of the 47 in total) \url{https://clinicaltrials.gov/ct2/results?term=Hydroxychloroquine&Search=Apply&recrs=d&age_v=&gndr=&type=&rslt=}; last accessed on 15-1-2021.}, so then the query ``retrieve all COVID-19 drugs'' will include in the query answer hydroxychloroquine, since it recursively retrieves the instances down in the class hierarchy for all {\sf\small COVID-19 drug} subclasses. Hydroxychloroquine is definitely not a drug to effectively treat COVID-19, however, nor has it been approved for that purpose in any country. 

None of the COVID-19 ontologies have any meaningful deductions along the line of the protein phosphatase experiment that deduced a novelty for human understanding of it at the theory level \cite{Wolstencroft07}, nor are they aimed at achieving that at present. Issues such as the robot in Example 1 and Fig.~\ref{fig:genetpart}, which similarly can be transposed on the gender binary bias, would likely surface during ontology development since typically the reasoner is used to eliminate errors and then the deductions are materialsed so that the reasoner is not needed in time-sensitive applications other than for query answering. Alternatively, a light-wight ontology language is used from the start so that disagreements do not surface due to lack of language expressiveness, notably because of the absence of disjointness and qualified cardinality constraints. Therefore, our expectation is that the effects of bias with respect to reasoning consequences may be more salient in data management and retrieving information rather than in reasoning over the logical theory itself.

\section{Conclusions}

Bias in the models easily creep into an ontology for various reasons. Eight types of sources of bias for ontologies were identified and illustrated: philosophical, purpose, science, granularity, linguistic, socio-cultural, political or religious, and economic motives. Some of them are explicit, and some may be either explicit or implicit. Three COVID-19 ontologies that were developed at the same time by different groups were assessed on these types of bias, which showed that each one exhibited a subset of the types of sources of bias. This first characterisation and comparative assessment may contribute to further research into ethical aspects of ontologies, both the modelling component and how it affects their use in applications.
 
As future work, we plan to look into a systematic way assessing and annotating explicit choices in the ontology, since  ontologies tend to be decoupled from any possible related ontology paper that otherwise could have provided context.


%
%
%
%

\end{document}